\documentclass{article}
\usepackage{spconf,amsmath,graphicx}
\usepackage{subcaption}
\usepackage{graphicx}
\usepackage{float}
 
\usepackage{multirow}
\usepackage{rotate}
\usepackage{indentfirst}
\usepackage[american]{babel}
\usepackage[ruled, linesnumbered]{algorithm2e}
\usepackage{microtype}
\usepackage{booktabs}
\usepackage{amssymb}
\usepackage{url}
\usepackage{bm}
\usepackage{amsfonts}
\usepackage{amsmath}
\usepackage{amsthm}
\usepackage{lipsum}
\usepackage{amstext}
\pdfoutput=1

\usepackage{hyperref}

\newcommand{\etal}{\textit{et al}.}
\newcommand{\ie}{\textit{i}.\textit{e}.}
\newcommand{\eg}{\textit{e}.\textit{g}.}

\title{Language-guided Few-shot Semantic Segmentation}
\name{Jing Wang$^{\star}$ \qquad Yuang Liu ${\dagger}$ \qquad Qiang Zhou $^{\star}$ \qquad Fan Wang$^{\star}$}
\address{$^{\star}$ DAMO Academy, Alibaba Group \\
      $^{\dagger}$ East China Normal University}

\begin{document}
%\ninept
%
\maketitle
\begin{abstract}
Few-shot learning is a promising way for reducing the label cost in new categories adaptation with the guidance of a small, well labeled support set. But for few-shot semantic segmentation, the pixel-level annotations of support images are still expensive. In this paper, we propose an innovative solution to tackle the challenge of few-shot semantic segmentation using only language information, \ie image-level text labels. Our approach involves a vision-language-driven mask distillation scheme, which contains a vision-language pretraining (VLP) model and a mask refiner, to generate high quality pseudo-semantic masks from text prompts. We additionally introduce a distributed prototype supervision method and complementary correlation matching module to guide the model in digging precise semantic relations among support and query images. The experiments on two benchmark datasets demonstrate that our method establishes a new baseline for language-guided few-shot semantic segmentation and achieves competitive results to recent vision-guided methods.
\end{abstract}
\begin{keywords}
Few-shot learning, semantic segmentation, vision-language
\end{keywords}
\section{Introduction}
\label{sec:intro}

Semantic segmentation is an important task of pattern recognition~\cite{mo2022review}, which aims to allocate a category label to each pixel. With the development of deep learning, the accuracy of semantic segmentation has risen dramatically, but with the growing need of large-scale dense labels. Meanwhile, the well-trained model cannot be directly applied to new categories until re-training. Few-shot learning is a recent trending topic who aims to solve the label shortage and quick adaptation problem in deep learning. Instead of training a task-specialized model from scratch, few-shot learning tries to train a task-independent model in a ``meta-learning'' paradigm to dig the common knowledge shared across different tasks~\cite{li2021concise}. The model can be easily adapted to new tasks with a few support samples after ``meta-training''. Many researchers have explored the efficiency of few-shot on classification~\cite{liu2022few,wertheimer2021few,snell2017prototypical}, object detection~\cite{antonelli2022few,kang2019few,zhang2021pnpdet}, and semantic segmentation~\cite{luo2022meta,wang2019panet,dong2018few}. Even the few-shot learning can decrease the cost of adaptation to new tasks, the ``meta-learning'' process requests a sufficient amount of well-labeled base data. Comparing to the image-level text label and the bounding box label, the pixel-wise dense segmentation map adopted in semantic segmentation is harder to acquire. In this paper, we consider a more valuable and challenging situation in few-shot semantic segmentation (FSS), \ie, language-guided few-shot semantic segmentation (LFSS), where only the image-level labels are available. 

\begin{figure}[t]
  \centering
  \subfloat[Regular FSS]{
    \includegraphics[width=0.48\linewidth]{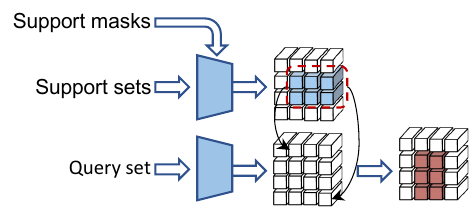}
  }
  \subfloat[Language-guided FSS]{
    \includegraphics[width=0.48\linewidth]{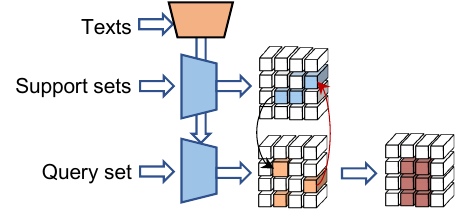}
  }
  \caption{Comparison between the regular few-shot segmentation (a) and the proposed language-guided approach (b). In regular few-shot, ground-truth of support masks are adopted to select representative features in support feature maps, and target features in query feature maps will be picked through a singe-direction matching. In the proposed method, the cheap but abstract text labels are adopted to mark target features in support and query images generally, then the double-direction matching can help to pick more accurate target features.}
  \label{fig:intro}
\end{figure}

The LFSS is rarely studied because of the information scarcity. Instead of dense masks, \cite{wang2019panet,zhang2019canet,Rakelly2018ConditionalNF} has explored to train the few-shot segmentation model by scribble, bounding box annotations, or sparse pixel annotations. These annotations are more sparse than the pixel-level annotations, but still require a strong artificial prior. ~\cite{raza2019weakly} firstly introduces class label supervision to FSS, they train the model following regular few-shot learning (fully-annotated support masks are necessary), but during testing, they only take the class labels as prior to lead the nearest neighbor classification and generate a general support proposal for object segmentation in query images. ~\cite{siam2020weakly} propose a novel multi-modal interaction module for few-shot segmentation, they design a co-attention mechanism to align the visual input and natural word embedding. To explore more information from the text labels, ~\cite{lee2022pixel} conduct the efficient classification activation maps (CAM)~\cite{selvaraju2017grad} to extract pseudo masks from category text labels as supervision. Due to the inaccurate pseudo masks and the gap between visual and text embedding, the performance of these language-guided works are far away from the vision-guided methods. 

Recently, ~\cite{wangiterative} has expanded the VLP model to few-shot learning, where they treat CLIP~\cite{radford2021learning} as an efficient classifier and conduct CAM to generate more accurate pseudo masks from text prompts. These pseudo masks directly take place of the ground-truth support masks to train the few-shot model. However, as pseudo masks can't be as subtle as the manual labels, training few-shot model in fully-supervised manner with them is suboptimal. In this paper, we propose a Language-guided Few-shot semantic Segmentation model (\textbf{LFSS}). It consists of a VLP-driven mask distillation \textbf{(VLMD)} mechanism for generating high quality pseudo masks and a custom feature learning module for digging exact guidance from coarse pseudo masks. Firstly, We employ MaskCLIP~\cite{zhou2022extract}, a semantic segmentation model expanded from CLIP~\cite{radford2021learning}, to transfer text labels into pseudo masks. We then adopt a mask refiner to remove false mask predictions. In vision-guided few-shot semantic segmentation, prototype learning is a widely adopted method where masked average pooling (MAP) extract one or few class prototypes from the regions of interest (ROI) in support feature maps. Matching support prototypes with query features can acquire the semantic similar target features~\cite{snell2017prototypical,tian2020prior,lang2022learning}. However, in LFSS, the coarse pseudo masks will lead to inaccurate prototypes. To address this, we have designed a distributed prototype supervision \textbf{(DPS)} and a complementary correlation matching \textbf{(CCM)} module to reduce the effect of the pseudo mask and reveal the correct semantic relations among the support and query images. 

% Our contribution can be summarized as follows:
% \begin{itemize}
%   \item We propose a language-guided framework to tackle the challenge of few-shot semantic segmentation, where the unseen objects can be segmented with only image-level text labels. 
  
%   \item By introducing a well-pretrained VLP model, \ie, MaskCLIP, the coarse pseudo masks can be distilled with only the prompt of text labels. And we design a mask refining module to calibrate the masks. 
  
%   \item To further prevent the interface of inaccurate mask, we propose a distributed prototype supervision (DPS) module to extract accurate semantic representatives under the context of coarse masks, and a complementary correlation matching (CCM) module to induce the model to dig correct semantic relation among support and query images.
  
%   \item The experimental results demonstrate that our method can not only surpass the state-of-the-art language-guided few-shot segmentation methods, but also achieves comparable performance to recent fully-supervised few-shot methods.
  
% \end{itemize}

\section{Related works}

\subsection{Few-shot Semantic Segmentation}
Many methods have been proposed to aggregate the guidance from support images to segment new objects of the same class in query images in few-shot style. For example, extracting representative prototypes from support feature maps by masked average pooling~\cite{snell2017prototypical,tian2020prior,lang2022learning}, calculating pixel-wise correlation between support and query features~\cite{min2021hypercorrelation,iqbal2022msanet}, and so on. However, pixel-wise annotation of support images are required for the regular few-shot segmentation models. To further reduce the label cost of training, language-guided methods are proposed in few-shot segmentation. ~\cite{wang2019panet,zhang2019canet,Rakelly2018ConditionalNF} tried to train the model with sparse labels like bounding box or scribbles. ~\cite{raza2019weakly} firstly proposed to train a regular few-shot segmentation model on base but testing it with only class labels. ~\cite{siam2020weakly} explored the effectiveness of combining the visual embedding with text embedding in few-shot segmentation. To reduce the gap between vision and text, ~\cite{lee2022pixel} extracted pseudo masks from text labels by CAM, and ~\cite{wangiterative} introducing the powerful vision-language pretraining model CLIP to transfer the text labels into pseudo masks and achieved comparable performance to the fully-supervised few-shot segmentation model.

\subsection{Vision-Language Model}

As a pioneering work towards vision-language pre-training, CLIP~\cite{radford2021learning} has promoted a wide range of multi-modal applications~\cite{ni2022expanding,gal2022stylegan,luo2022clip4clip} and shows great potential in zero-shot or few-shot vision tasks~\cite{gu2021open,ding2022decoupling,pourpanah2022review}. Especially, a group of researchers has extended it into dense prediction tasks, \eg, semantic segmentation~\cite{rao2022denseclip,ding2022decoupling}, image generation~\cite{gal2022stylegan} and object detection~\cite{gu2021open}. DenseCLIP~\cite{rao2022denseclip} is the pioneer that employs CLIP in semantic segmentation and tickles the issue of pixel-text matching via context-aware prompting. Ding~\etal~\cite{ding2022decoupling} decouples the zero-shot segmentation task as a class-agnostic grouping task and a zero-shot classification task to perform segment-text matching. However, the above methods all depend on complicated prompt engineering, and are limited to the lack of fine-annotated images.

\begin{figure}[!t]
 \centering
 \includegraphics[width=\linewidth]{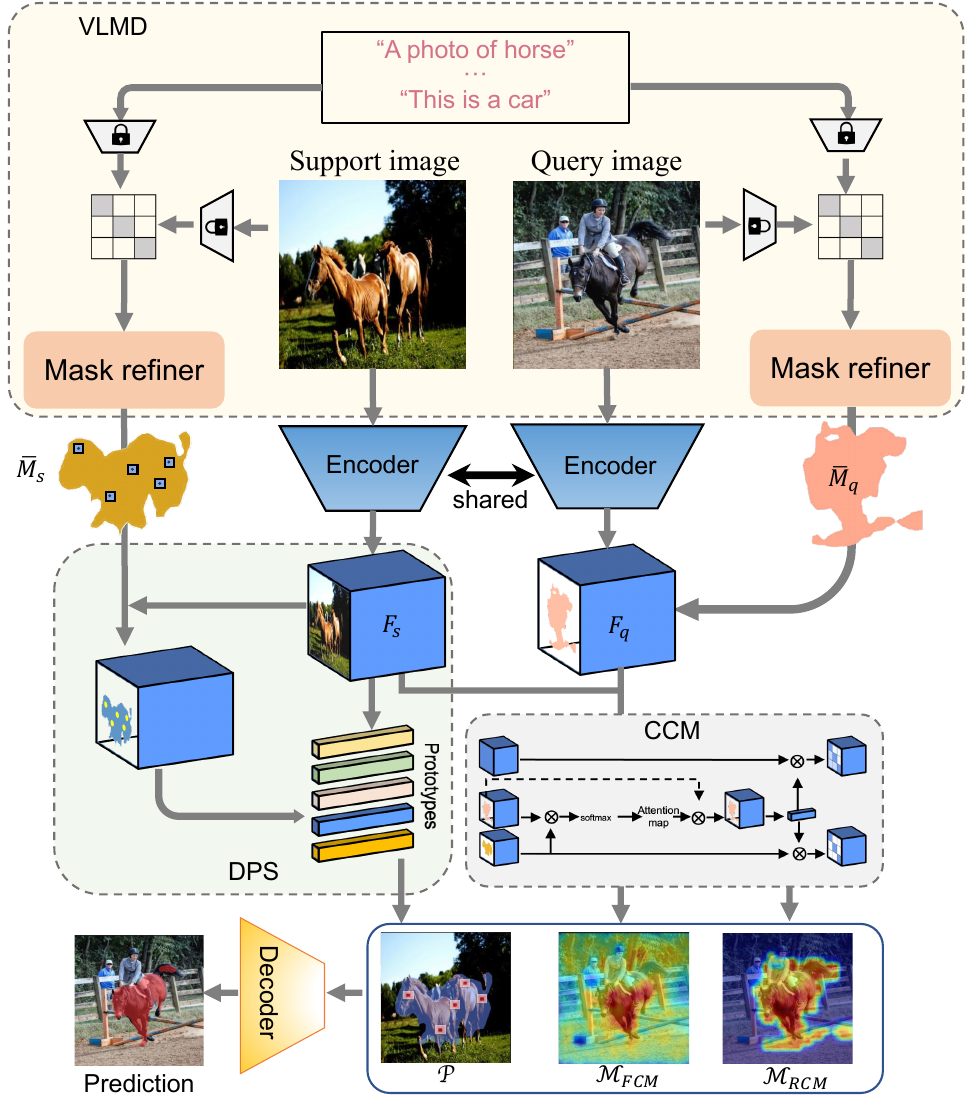}
 \caption{Overview of our LFSS framework, which consists of the proposed vision language pre-training model-driven mask distillation (VLMD), distributed prototype supervision module (DPS), and complementary correlation matching module (CCM).}
 \label{fig:framework}
\end{figure}

\section{Method}

\subsection{Problem setup}

 For a regular 1-way K-shot few-shot segmentation task $T$, a support set $S={(I_s,M_s)}$ and a query set $Q={(I_q,M_q)}$ are required, where $I$ and $M$ represent image and ground truth mask respectively, $|S| = K$, $S$ and $Q$ are sampled from the same category. The goal is to train model who can predict $M_q$ for $I_q$ with a given $S$, subject to $K$ being small for few-shots. In this paper, we consider a more challenging setting in few-shot semantic segmentation, where only the text class labels ($L$) of the support images are available, \ie $S={(I_s,L_s)}$ and $Q={(I_q,L_q)}$. We adopt the widely used episodic training paradigm to train our model, where datasets are split into $D_{train}$ and $D_{test}$ with category set $C_{train}$ and $C_{test}$ respectively, and $C_{train} \bigcap C_{test} = \phi$. We repeatedly sample task $T$ from $D_{train}$ during training, and the trained model are directly evaluated on $D_{test}$ to predict $M_q$ for $I_q$ \ie:

\begin{equation}
  \hat{M_q} = f(\{(I_s^k , L_s^k)\}_{k=1}^K , I_q,\theta \| c \in C_{test})
\end{equation}
where $f(, \| \theta)$ is the trained model.

\subsection{Overview}

In this work, we aim to train an accurate few-shot semantic segmentation model with only text labels. The overall architecture of the proposed method are shown as Figure~\ref{fig:framework}, which is a double-branch architecture, consisted of the vision language pre-training driven mask distillation module (VLMD) and a custom feature learning stream. The support and query images along with language descriptions are first fed to the VLMD to extract reliable pseudo masks. At the same time, the backbone will extract multi-level features from support and query images respectively. Then with the guidance of the pseudo masks, a distributed prototype supervision (DPS) module are applied on the support features to extract local representative prototypes and a complementary correlation matching (CCM) module learns to generate a fine-grained correlation map by matching the query and support features. We will take the calculation process of one-shot as example to introduce these effective modules amply in the follow sections.

\subsection{VLP-driven Mask Distillation (VLMD)}
 As text labels are abstract and information-limited, acquiring more information from them poses the first challenge. To tackle this, we introduce a VLMD module to project the text labels to pseudo masks, which consisting of a mask generator and mask refiner. For segmenting targets annotated by text labels in images, we adopts the VLP model, MaskCLIP ~\cite{zhou2022extract}, to generate high quality pseudo masks. Specifically, we adopt the modified ResNet as image encoder, then we remove the query and key embedding layers from the last global attention pooling layer, and directly feed the feature map from the final residual block into the value-embedding layer and the following linear layer, which are reformulated into two respective 1 $\times$ 1 convolution layers to keep the spatial dimension of feature maps (this process can be visualized in Fig.2(b) of ~\cite{zhou2022extract}). The text encoder are unchanged. The cosine similarity between the text embedding and the image feature maps can tell the category of each pixel.

\begin{figure}[!t]
 \centering
 \includegraphics[width=\linewidth]{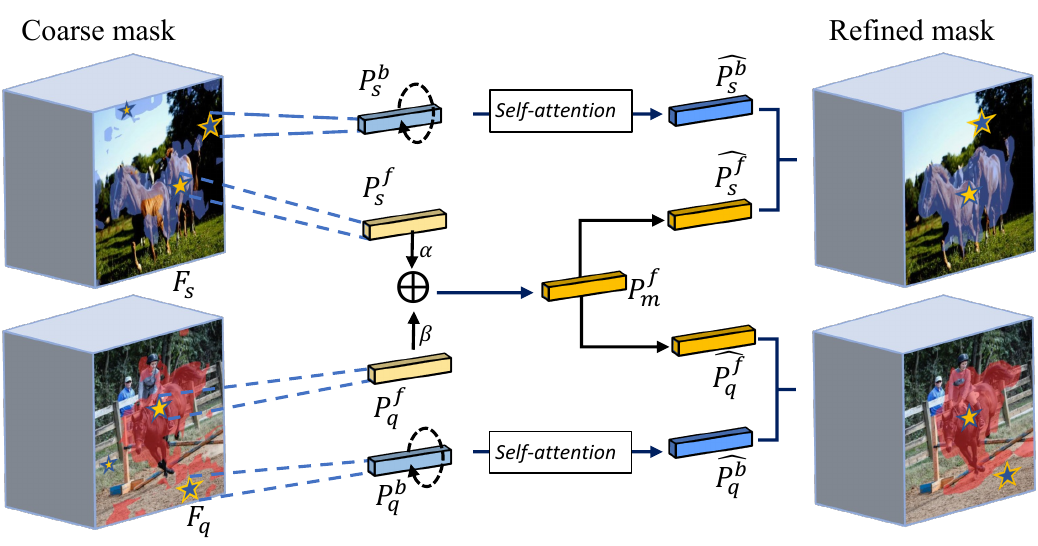}
 \caption{The detail of self-supported mask refinement module. }
 \label{fig:corr}
\end{figure}

However, despite the MaskCLIP model can help to generate high quality pseudo masks, they can not as elaborate as the manual masks used in regular few-shot segmentation. To reduce false predictions in these pseudo masks, we introduce a mask refiner to improve their accuracy.Leveraging the notion that pixels belonging to the same object are more similar than those to different objects of same class, we adopt a self-supported approach to refine the initial pseudo masks. As illustrated in Figure~\ref{fig:corr}, features extracted from the backbone can be separated into foreground and background features based on the initial pseudo masks, then we conduct MAP to aggregate the respective foreground prototypes and background prototypes from support and query feature maps:
\begin{equation}
  P^f = \frac{\sum_{x=1,y=1}^{w,h} F_{x,y} \odot M_{x,y}}{\sum_{x=1,y=1}^{w,h} M_{x,y}} \,,
\end{equation}

where $F \in R^{c \times w \times h}$ represents features that extracted by backbone network, $\odot$ is Hadamard product. As the query and support images should contain objects of the same class, we add the foreground prototypes to weight the specified objects, formulated as follows:
\begin{equation}
  P_m^f = \alpha P_s^f + (1 - \alpha) P_q^f \,,
\end{equation}
$\alpha$ is the balance factor, $s$ and $q$ represent support and query set respectively. 
Different from the foreground, the background prototypes are independently responsible for corresponding feature maps as the background between support and query sets are quit different, and a self-attention operation is adopted to acquire the background prototypes:

\begin{equation}
  P^b = \mathtt{softmax}(F_b \cdot F^\top) \cdot F_b \,,
\end{equation}
where $F_b=F\odot (1-M)$represents the background features, $F$ represents the full feature map. Then we calculate the cosine similarity between the features and prototypes to obtain new masks:
\begin{equation}
    S^f = \frac{F \cdot P^\top}{\left\lVert F \right\rVert \left\lVert P \right\rVert } \,,
\end{equation}
where $P \in \{P_m^f,P_b\}$.
Then the features are assigned to foreground or background according to the similarity score. After mask refinement, most false predictions of the initial masks can be removed, acquired the refined support mask $\bar{M}_s$ and query mask $\bar{M}_q$. 

\subsection{Distributed Prototype Supervision (DPS)}

\begin{figure}[!t]
 \centering
 \includegraphics[width=\linewidth]{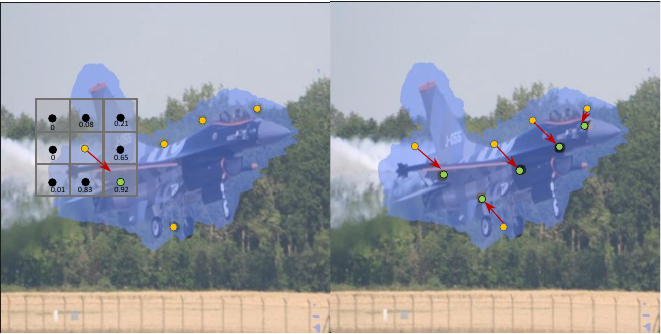}
 \caption{The detail of selecting the super-pixel seeds. }
 \label{fig:DPS}
\end{figure}

Prototype learning is a popular feature alignment method in few-shot segmentation~\cite{shaban2017one,dong2018few,lang2022learning}. Typically, all foreground support features are compressed into a global prototype by MAP (refer to Eq.(1)), which is semantically rich but lack of spatial information. To solve this issue, we designed a custom Distributed Prototype Supervision (DPS) module, which extracts multiple local prototypes from the coarse pseudo masks instead of a global prototype. As shown in Figure~\ref{fig:DPS}, we first distribute $N_{sp}$ initial seed points in the pseudo mask, where a Euclidean distance transform is adopted to place the seed points far from the boundary of mask and other seed points:
\begin{equation}
    D(x,y) = \min_{l \in L} \sqrt{((x-x_l)^2 + (y-y_l)^2)}\,,
\end{equation}
where $x$ and $y$ represent the spatial coordinate values of a seed, the max $D(x,y)$ represents the furthest distance. $L=B \cup P$ represents the background feature points($B$) and the labeled points ($P$), the selected points will be added to $P$ after each iteration.
After placing the initial seed points, we extract corresponding features in feature map as super-pixel seeds:$O^0 \in \{R^{C \times N_{sp}}\}$ ($C$ is the channels of feature map). To prevent the incorrect placement of seed points in the background region of pseudo coarse mask, we utilize a part-aware module to rectify the location of the initial seed points. As shown in Fig~\ref{fig:DPS}, after placing a seed point, we sample an $n * n$ grid, \ie $G$, around it and calculate the similarity between features locate in $G$ and the $P_q^f$: 
\begin{equation}
    S_{i,j} = \frac{g_{i,j} \cdot (P_q^f)^\top}{\left\lVert g_{i,j} \right\rVert_2 \left\lVert P_q^f \right\rVert_2 } \,,
\end{equation}
where $S_{i,j}$ represents the similarity score, $g_{i,j} \in \mathbb{R}^{1\times C}$ means support features locate at $(i,j)$ in $G$.
The point whose corresponding feature with the highest similarity score will replace the original seed, formulated as follows:
\begin{equation}
    \hat{i},\hat{j} =\mathtt{argmax}_{i,j}(S_{i,j})\,.
\end{equation}
After adjusting the seed points, we assume that all seeds are located at target objects and extract the new super-pixel seeds $O^0 \in \{R^{C \times N_{sp}}\}$. To extract semantic prototypes, we cluster the feature map into $N_{sp}$ super-pixel with guidance of the super-pixel seeds. We firstly add coordinates of each pixel to the feature maps to increase spatial priors. Then we cluster feature maps in an iterative manner. During each iteration, we first calculate the correlation map $C^t$ between each foreground feature point $p$ and all super-pixel seeds:
\begin{equation}
    C_{p,i}^t = e^{-Q(F_p,O_i^{t-1})}\,,
\end{equation}
where $F_p$ represents foreground pixels, $i \in N_{sp}$. $Q$ is a distance function defined as:
\begin{equation}
    Q(F,O) = \sqrt{\left(d_f(F_1,F_2)\right)^2 + \left(\frac{d_s(O_1,O_2)}{r}\right)^2} 
\end{equation}
where $d_f$ and $d_s$ are Euclidean distance for features and coordinate values, $r$ is a temperature value~\cite{li2021adaptive}. 
Then we update the super-pixel centroids following:
\begin{equation}
    O_i^t = \frac{1}{\sum_{N_{fp}} C_{p,i}^t}\sum_{p=1}^{N_{fp}} C_{p,i}^t F_{p}\,,
\end{equation}
where $N_{fp}$ is the foreground pixels number. After clustering, the resulting super-pixel centroids are treated as the part-aware prototypes, dubbed as $P_{sc}$. Instead of expanding the prototypes to specified shape and concatenating them with feature maps, we calculate association map between the $P_{sc}$ and support feature map instead:
\begin{equation}
    \mathcal{P} = \sum_i^{N_{sp}} \frac{P_{sc} \cdot F_s}{\left\rVert P_{sc}\right\rVert \left\rVert F_s\right\rVert}\,.
\end{equation}

\subsection{Complementary Correlation Matching (CCM)}

Even prototypes work effectively in matching objects with semantic similarity, but the sparse nature stops them from fine-grained relation exploitation. To make better use of the pseudo mask, we proposed a complementary correlation matching module (CCM), which consisted of a ROI-guided correlation matching (RCM) and a full image correlation matching (FCM). We first extract an attention map from the query image and support image with the guidance of their pseudo masks, formulated as follows:
\begin{equation}
    \mathcal{A} = \mathtt{softmax}\left(\frac{(F_q \odot \bar{M_q}) \cdot {(F_s \odot \bar{M_s})^\top}}{\left\rVert F_q \odot \bar{M_q} \right\rVert \left\rVert F_s \odot \bar{M_s} \right\rVert}\right) \,.
\end{equation}
As the most common part of the query and support images should be the objects of the specified class, we highlight the target area by multiplying the support feature maps with the attention map $\mathcal{A}$ and extracting a more focused prototype $P_a$ by MAP:
\begin{equation}
    P_a = \frac{\sum_{x=1,y=1}^{w,h} \mathcal{A}^{x,y} (F_s^{x,y}  \bar{M}_s^{x,y})}{\sum_{x=1,y=1}^{w,h} \bar{M}_s^{x,y} }\,.
\end{equation}
Then we obtain the ROI-guided correlation map by matching $P_a$ with the masked query feature map:
\begin{equation}
    \mathcal{M}_{RCM} = \frac{P_a \cdot (F_q \odot \bar{M_q})^\top}{\left\rVert P_a\right\rVert \left\rVert F_q \odot \bar{M_q}\right\rVert}\,.
\end{equation}

The $\mathcal{M}_{RCM}$ helps locate exact objects in query image from the coarse masked ROI, however, it's isolated from those omitted by the pseudo masks. To solve this problem, we further extract the FCM by matching all query features with the $P_a$:
\begin{equation}
    \mathcal{M}_{FCM} = \frac{P_a \cdot F_q^\top }{\left\rVert P_a\right\rVert \left\rVert F_q \right\rVert}\,.
\end{equation}
The RCM and FCM works complementarily to detect all targets in query images, so we concatenate them together to get the fine-grained correlation map:
\begin{equation}
    \mathcal{M} = \mathcal{M}_{RCM} \oplus \mathcal{M}_{FCM}\,,
\end{equation}
where $\oplus$ is the channel-wise concatenation operation.
Finally, we concatenate the query features $F_q$ with prototype-associated map $\mathcal{P}$ and the fine-grained correlation map $\mathcal{M}$ to obtain more guidance, thus the final feature map $\mathcal{F}$ that fed to the decoder is:
\begin{equation}
    \mathcal{F} = F_q \oplus \mathcal{P} \oplus \mathcal{M}\,.
\end{equation}
 The final prediction is acquired by:
 \begin{equation}
     \hat{M_q} = \textbf{Dec}(\mathcal{F})\,,
 \end{equation}
 \textbf{Dec} is a light-weight decoder.

\subsection{Objective Function}

The binary cross entropy (BCE) loss is adopted to train the model. To speed up convergence, we employ a circle training strategy. Specifically, the support image is firstly deemed as query image and fed to the model to acquire an $\hat{M_s}$. Then $\hat{M_s}$ is set as the new support mask to support the prediction of query mask $\hat{M_q}$. The overall loss function is formulated as: 
\begin{equation}
    \mathcal{L} = \beta \mathcal{L}_{BCE}(\hat{M_s},M_s^{gt}) + (1 - \beta) \mathcal{L}_{BCE} (\hat{M_q},M_q^{gt}) \,,
\end{equation}
where $M_s^{gt}$/$M_q^{gt}$ represent the ground-truth of support/query sample, $\beta$ is the balance factor.

\section{Experiments}

\subsection{Experimental Settings}

We evaluate our approach on two public datasets that widely enrolled in regular few-shot semantic segmentation, \ie, Pascal-$5^i$~\cite{shaban2017one} and COCO-$20^i$~\cite{lin2014microsoft} ($i$ is the number of folds). Following the setting of few-shot segmentation, we split each dataset into four folds, set three of them as training set and sample 1000 episodes from the remaining fold as test set. The mean intersection over union (mIoU) of all classes is utilized to measure the performance. To fairly compare with state-of-the-art (SOTA) methods, we set the popular convolution neural network VGG-16 and ResNet-50 pretrained on ImageNet as backbone, a light-weight decoder contains an ASPP (atrous spatial pyramid pooling) block and three plain convolution blocks works for the final segmentation. The pretrained MaskCLIP is adopted for initial pseudo masks generation, the visual and text encoders of MaskCLIP are modified ResNet-50. The backbone and MaskCLIP are frozen during model training to prevent overfitting.

For mask generation, we expand the text label with 85 prompt templates followed MaskCLIP~\cite{zhou2022extract} and fed them to the text encoder, then average the processed text embeddings of the same class. We resize the input to 400 $\times$ 400 in both training and testing stage following~\cite{wangiterative}, and no extra data augmentation trick is adopted. The learning rate is set to 0.001. We train the model for 200 epochs on 8 NVIDIA V100 GPUs with Adam optimizer. The hyperparameters $\alpha$ and $\beta$ are empirically set at 0.5, and $n=3$ for saving calculation.

\begin{table}[!t]
  \centering
  \renewcommand{\arraystretch}{1.1} % Default value: 1
  \caption{Comparisons with fully-supervised FSS and LFSS methods on Pascal-$5^i$. ``P'', ``I'' and ``B'' represent the three types of semantic annotation (``Ann.''): Pixel, Image and Box. ``BB.'' means the backbone. The ``-'' is placeholder for unreported results by original paper.}
  \resizebox{\linewidth}{!}{
    \begin{tabular}{c|lc|cccc|c}
    \toprule
    BB. & \multicolumn{1}{l}{Method} & Ann. & $5^0$  & $5^1$  & $5^2$  & $5^3$  & Mean \\
    \midrule
    \multirow{6}{*}{\rotatebox{270}{VGG-16}} & PFENet & P & 56.9  & 68.2  & 54.4  & 52.4  & 58.0 \\
          & HSNet & P & 59.6  & 65.7  & 59.6  & 54.0  & 59.7 \\
\cline{2-8}          & PANet & B   & -     & -     & -     & -     & 45.1 \\
          & CAM-WFSS & I & 36.5  & 51.7  & 45.9  & 35.6  & 42.4 \\
          & IMR-HSNet & I & \textbf{58.2} & 63.9  & 52.9  & 51.2  & 56.5 \\
          & \textbf{Ours} & I & 56.3  & \textbf{65.2} & \textbf{53.6} & \textbf{55.7} & \textbf{57.7} \\
    \midrule
    \multirow{7}{*}{\rotatebox{270}{ResNet-50}} & PFENet & P & 61.7  & 69.5  & 55.4  & 56.3  & 60.8 \\
          & HSNet & P & 64.3  & 70.7  & 60.3  & 60.5  & 64.0 \\
          & ASGNet & P & 58.8  & 67.9  & 56.8  & 53.7  & 59.3 \\
\cline{2-8}          & CANet & B   & -     & -     & -     & -     & 52.0 \\
          & VS-WFSS & I & 42.5 & 64.8 & 48.1 & 46.5 & 50.5 \\
          & IMR-HSNet & I & \textbf{62.6} & \textbf{69.1}  & 56.1  & 56.7  & 61.1 \\
          & \textbf{Ours} & I & 59.9  & \textbf{69.1} & \textbf{56.7} & \textbf{58.9} & \textbf{61.2} \\
    \bottomrule
    \end{tabular}%
  }
  \label{tab:voc}%
\end{table}%

\begin{table}[t]
  \centering
  \renewcommand{\arraystretch}{1.1}
  \caption{Comparisons with fully-supervised FSS and LFSS methods on COCO-$20^i$. }
  \resizebox{\linewidth}{!}{
    \begin{tabular}{c|lc|cccc|c}
    \toprule
    BB. & \multicolumn{1}{l}{Method} & Ann. & $20^0$ & $20^1$ & $20^2$ & $20^3$ & Mean \\
    \midrule
    \multirow{6}{*}{\rotatebox{270}{VGG-16}} & PFENet & P & 35.4  & 38.1  & 36.8  & 34.7  & 36.3 \\
          & BAM-base & P & 39.0  & 47.0  & 46.4  & 41.6  & 43.5 \\
\cline{2-8}          & PANet & B   & 12.7  & 8.7   & 5.9   & 4.8   & 8.0 \\
          & CAM-WFSS & I & 24.2  & 12.9  & 17.0  & 14.0  & 17.0 \\
          & IMR-HSNet & I & 34.9  & 38.8  & 37.0  & 40.1  & 37.7 \\
          & \textbf{Ours} & I & \textbf{37.6} & \textbf{49.6} & \textbf{42.5} & \textbf{43.4} & \textbf{43.3} \\
    \midrule
    \multirow{7}{*}{\rotatebox{270}{ResNet-50}} & PFENet & P & 36.5  & 38.6  & 34.5  & 33.8  & 35.8 \\
          & HSNet & P & 36.3  & 43.1  & 38.7  & 38.7  & 39.2 \\
          & BAM-base & P & 41.9  & 45.4  & 43.9  & 41.2  & 43.1 \\
          & ASGNet & P & -     & -     & -     & -     & 34.6 \\
\cline{2-8}          & VS-WFSS & I & -     & -     & -     & -     & 15.0 \\
          & IMR-HSNet & I & 39.5  & 43.8  & 42.4  & 44.1  & 42.4 \\
          & \textbf{Ours} & I & \textbf{42.9} & \textbf{51.8} & \textbf{44.4} & \textbf{46.8} & \textbf{46.4} \\
    \bottomrule
    \end{tabular}%
  }
  \label{tab:coco}%
\end{table}%

\begin{figure*}[!t]
 \centering
 \includegraphics[width=1.0\linewidth]{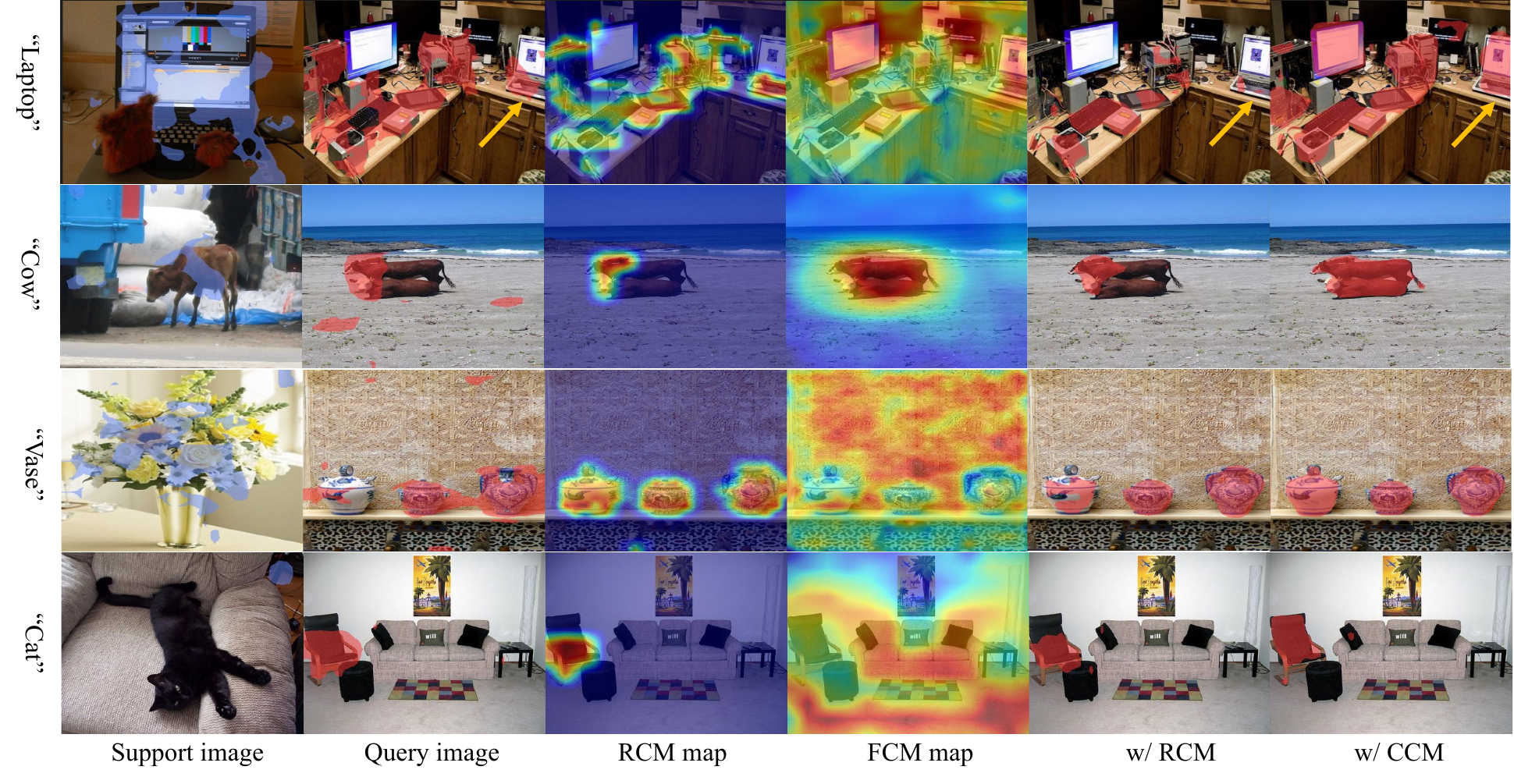}
 \vspace{-3mm}
 \caption{Visualization of RCM (the 3-rd column), FCM (the 4-th column) map and the corresponding segmentation results with RCM and CCM respectively (the last two columns). }
 \label{fig:CCM}
\end{figure*}

\subsection{Comparison with State-Of-The-Art}

    We compared our methods with the SOTA LFSS, \ie CAM-WFSS~\cite{lee2022pixel}, IMR-HSNet~\cite{wangiterative}, VS-WFSS~\cite{siam2020weakly}, and some recent fully-supervised few-shot segmentation (FFS) works, \ie PFENet~\cite{tian2020prior}, PANet~\cite{wang2019panet}, HSNet~\cite{min2021hypercorrelation},  ASGNet~\cite{li2021adaptive}, BAM-base~\cite{lang2022learning}, in 1-shot setting. The results on Pascal-$5^i$ are displayed in Table ~\ref{tab:voc}. With only text supervision, all our model with different backbones surpass other LFSS methods and most FSS methods. For FSS, our method surpasses the prototype-based PFENet but is not as excellent as the HSNet, who introduced pixel-level correlation to achieve fine-grained feature alignment. Specially, the proposed method outperforms the language-guided version HSNet, \ie, IMR-HSNet. Our model exceeds the IMR-HSNet with 1.2\% and 0.1\% mIoU for VGG-16 and resnet-50 backbones respectively. The IMR-HSNet directly adopts HSNet to train the LFSS model but neglects that the gap between the elaborate manual labels and the coarse pseudo masks. Take this in mind, we design this custom network to mitigate the effect of false predictions in pseudo masks and achieve better results.

    Table~\ref{tab:coco} summaries the evaluation results on COCO-$20^i$, which is a more challenging dataset contains 80 categories, many FSS models' performance dropped on this dataset because of its complexity. However, our method shows excellent generality on COCO, exceed SOTA LFSS method, \ie IMR-HSNet, by a large margin (5.6\% with VGG-16 and 4.0 \% with ResNet-50). False predictions of pseudo masks are more general in COCO dataset due to its variety. The proposed model designs the VLMD to generate high quality masks and reduce apparent errors, followed by custom DPS and CCM who learn to dig exact information from the pseudo masks to provide more guidance for targets segmentation. As a result, we not only outperform the LFSS, but also surpass recent FSS, \ie, BAM and HSNet. Our method with ResNet-50 backbone improves 7.2\% mIoU over the HSNet on COCO, but lost behind it on Pascal dataset. We infer that the pixel-level correlation proposed by HSNet is not good at extracting key information from complex scenario. As data in COCO is category-diverse and appearance-diverse, the pixel-level correlation is harder to dig and easy to be disturbed by other objects. In our method, the pseudo masks will locate the targets generally and guide the prototype extraction and correlation matching, which eases the few-shot training.

\begin{figure*}[!t]
 \centering
 \includegraphics[width=1.0\linewidth]{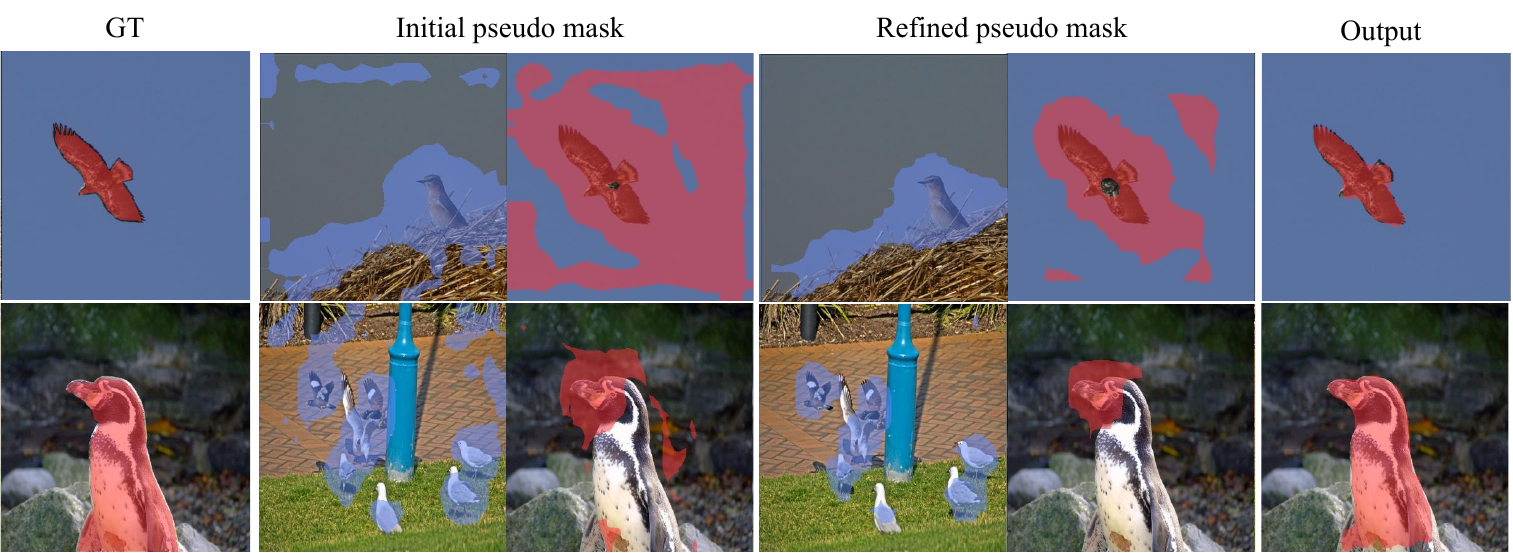}
 \caption{Qualitative results of initial mask, refined mask and output mask.}
 \label{fig:Segmentation}
\end{figure*}

% Table generated by Excel2LaTeX from sheet 'Sheet7'
\begin{table}[t]
  \centering
  \caption{The mIoU performance between pseudo masks (initial and refined) and the ground-truth.}
    \begin{tabular}{c|cc}
    \toprule
    Dataset & Initial mask & Refined mask \\
    \midrule
    VOC   & 26.94 & 32.52 \\
    COCO  & 26.99 & 33.84 \\
    \bottomrule
    \end{tabular}%
  \label{tab:maskrefine}%
\end{table}%

\subsection{Ablation Study}

Ablation studies are conducted to excavate the effectiveness of each component. We first evaluate the VLMD, Table~\ref{tab:maskrefine} depicts the mIoU between the pseudo mask and its corresponding ground-truth, we find the pseudo masks become more concise after refinement, the mIoU improves 5.58\% and 6.85\% on Pascal-$5^{i}$ and COCO-$20^i$ respectively. 

For feature learning module, we adopt ResNet-50 as feature extractor, and set a simple baseline by concatenating the global prototype and RCM as feature map. All models contain the same decoder, and the concatenated feature map is fed to the decoder directly to segment objects. The mean IoU on all categories of Pascal and COCO are summarized in Table~\ref{tab:ablation}.  Effected by coarse mask, the global prototype contains part of background information, so the results of baseline model are just passable. To improve the effectiveness of the prototype, we firstly replace the global prototype with DPS module to induce the model to focus on specific objects part instead of background, the model performs better on COCO dataset but worse on Pascal according to the results. We find the pseudo masks of Pascal data contain more false predictions as Pascal contains fewer categories while the MaskCLIP tries to annotate every pixel to a category. To curb the effect of false positives in pseudo masks, we distill more accurate masks by self-supported mask refiner. With finer masks, the DPS can extract valuable prototypes from target objects and the RCM can extract more focused association map. Quantitatively, the model's performance improves a lot after received finer masks (5\% on Pascal and 2.3\% on COCO). Finally, we introduce the CCM to capture more target information and prevent the omission of target in query images. The results are further improved on two dataset (2.4\% on Pascal and 6.7\% on COCO). 
\begin{table}[t]
  \centering
  \caption{Effectiveness of each component in our LFSS framework.}
    \begin{tabular}{c|ccc|c}
    \toprule
          Dataset   & DPS    & Mask refine & CCM   & mIoU (\%) \\
    \midrule
    \multirow{4}{*}{VOC} &       &       &       & 53.9 (baseline) \\
          & \checkmark     &       &       & 53.6 ($\downarrow$ 0.3) \\
          & \checkmark    & \checkmark     &       & 58.6 ($\uparrow$ 4.7) \\
          & \checkmark     & \checkmark    & \checkmark     & \textbf{61.2} ($\uparrow$7.3) \\
    \midrule
    \multirow{4}{*}{COCO} &       &       &       & 36.1 (baseline) \\
          & \checkmark     &       &       & 37.4 ($\uparrow$1.3) \\
          & \checkmark     & \checkmark     &       & 39.7 ($\uparrow$3.6) \\
          & \checkmark     & \checkmark     & \checkmark   & \textbf{46.4} ($\uparrow$10.3) \\
    \bottomrule
    \end{tabular}%
  \label{tab:ablation}%
\end{table}%

Moreover, we implement extra studies on VOC-$5^1$ to find out suitable hyperparameters ($\alpha$, $n$) for our model, the results are displayed in Tab~\ref{tab:hyperparamter}.  It's found that the model achieves best performance when $\alpha=0.5$ and $n=3$.

\begin{table}[h]
\centering
\caption{Impacts of $n$ and $\alpha$ on first fold of VOC-$5^i$. We set $\alpha$=0.5 for testing $n$ and set $n$=3 in reverse.}
\label{tab:hyperparamter}
\resizebox{.49\columnwidth}{!}{
\begin{subtable}[t]{0.495\linewidth}
    \begin{tabular}{c|c|c|c}
    \hline
    $n$     & 1     & 3     & 5 \\
    \hline
    mIoU  & 68.0    & \textbf{69.1}  & 68.7 \\
    \hline
    \end{tabular}%
    % \caption{$n$}
\end{subtable}
}
\resizebox{.49\columnwidth}{!}{
\begin{subtable}[t]{0.495\linewidth}
    \begin{tabular}{c|c|c|c}
    \hline
    $\alpha$ & 0.1   & 0.3   & 0.5 \\
    \hline
    mIoU  &  68.2    &  68.6 & \textbf{69.1} \\
    \hline
    \end{tabular}%
    % \caption{$\alpha$}
\end{subtable}
}
\end{table}

\subsection{Visualization}

To observe the results more intuitively, we visualized the association map generated by CCM, the refined masks, and some segmentation results respectively. As shown in Figure~\ref{fig:CCM}, the first two rows display samples that MaskCLIP failed to detect target objects in query images (annotated by yellow arrows), we found that the RCM also omitted these targets effected by the pseudo masks. So the model fail to segment them when only RCM is included (w/ RCM). Fortunately, we found the FCM will help to relocate the omitted objects after a full image matching. The last two rows display samples that with sick quality support masks, we find the RCM works effectively as targets in query images are detected by pseudo masks. Misrecognition and omission of targets are common during mask generation as we directly applied the general MaskCLIP to segment Pascal and COCO without fine-tuning. To this end, we add the $\mathcal{M}_{RCM}$ and $\mathcal{M}_{FCM}$ to acquire the final CCM that contains all possible target objects to improve model's performance.

The qualitative results of segmentation are plotted in Figure~\ref{fig:Segmentation}. The initial pseudo masks from MaskCLIP are coarse who contain many false positives (the second column, support images with light blue mask and query image with light red mask). The proposed mask refiner works effectively in reducing the wrongly recognized background (the third column). Even the refined masks are still rough and might omit some target areas, our method can induce the model to focus on exact target and achieve accurate segmentation (the final column).

\section{Conclusion}

In this work, we have tackled the challenge of languge-guided semantic segmentation by introducing a pretrained VLP model to generate pseudo masks from text labels as full-supervision. To reduce the false positives of pseudo masks and mine pure foreground representation, we propose a mask refine algorithm and a distributed prototype supervision strategy. The complementary correlation matching module learns a comprehensive fine-grained attention map to avoid objects omission. The extensive experiments on two public datasets evaluate the outstanding performance of our method, and the ablation study demonstrates the effectiveness of each component. In the future work, we plan to explore more complex LFSS tasks like general few-shot semantic segmentation by distilling more information from vision-language models. 
\bibliographystyle{IEEEbib}
\bibliography{re_refs}

\begin{thebibliography}{10}

\bibitem{mo2022review}
Yujian Mo, Yan Wu, Xinneng Yang, Feilin Liu, and Yujun Liao,
\newblock ``Review the state-of-the-art technologies of semantic segmentation based on deep learning,''
\newblock {\em Neurocomputing}, vol. 493, pp. 626--646, 2022.

\bibitem{li2021concise}
Xiaoxu Li, Zhuo Sun, Jing-Hao Xue, and Zhanyu Ma,
\newblock ``A concise review of recent few-shot meta-learning methods,''
\newblock {\em Neurocomputing}, vol. 456, pp. 463--468, 2021.

\bibitem{liu2022few}
Ying Liu, Hengchang Zhang, Weidong Zhang, Guojun Lu, Qi~Tian, and Nam Ling,
\newblock ``Few-shot image classification: Current status and research trends,''
\newblock {\em Electronics}, p. 1752, 2022.

\bibitem{wertheimer2021few}
Davis Wertheimer, Luming Tang, and Bharath Hariharan,
\newblock ``Few-shot classification with feature map reconstruction networks,''
\newblock in {\em CVPR}, 2021, pp. 8012--8021.

\bibitem{snell2017prototypical}
Jake Snell, Kevin Swersky, and Richard Zemel,
\newblock ``Prototypical networks for few-shot learning,''
\newblock {\em Advances in neural information processing systems}, vol. 30, 2017.

\bibitem{antonelli2022few}
Simone Antonelli, Danilo Avola, Luigi Cinque, Donato Crisostomi, Gian~Luca Foresti, Fabio Galasso, Marco~Raoul Marini, Alessio Mecca, and Daniele Pannone,
\newblock ``Few-shot object detection: A survey,''
\newblock {\em ACM Computing Surveys (CSUR)}, pp. 1--37, 2022.

\bibitem{kang2019few}
Bingyi Kang, Zhuang Liu, Xin Wang, Fisher Yu, Jiashi Feng, and Trevor Darrell,
\newblock ``Few-shot object detection via feature reweighting,''
\newblock in {\em ICCV}, 2019, pp. 8420--8429.

\bibitem{zhang2021pnpdet}
Gongjie Zhang, Kaiwen Cui, Rongliang Wu, Shijian Lu, and Yonghong Tian,
\newblock ``Pnpdet: Efficient few-shot detection without forgetting via plug-and-play sub-networks,''
\newblock in {\em WACV}, 2021, pp. 3823--3832.

\bibitem{luo2022meta}
Shuai Luo, Yujie Li, Pengxiang Gao, Yichuan Wang, and Seiichi Serikawa,
\newblock ``Meta-seg: A survey of meta-learning for image segmentation,''
\newblock {\em Pattern Recognition}, p. 108586, 2022.

\bibitem{wang2019panet}
Kaixin Wang, Jun~Hao Liew, Yingtian Zou, Daquan Zhou, and Jiashi Feng,
\newblock ``Panet: Few-shot image semantic segmentation with prototype alignment,''
\newblock in {\em ICCV}, 2019, pp. 9197--9206.

\bibitem{dong2018few}
Nanqing Dong and Eric~P Xing,
\newblock ``Few-shot semantic segmentation with prototype learning.,''
\newblock in {\em BMVC}, 2018.

\bibitem{zhang2019canet}
Chi Zhang, Guosheng Lin, Fayao Liu, Rui Yao, and Chunhua Shen,
\newblock ``Canet: Class-agnostic segmentation networks with iterative refinement and attentive few-shot learning,''
\newblock in {\em CVPR}, 2019, pp. 5217--5226.

\bibitem{Rakelly2018ConditionalNF}
Kate Rakelly, Evan Shelhamer, Trevor Darrell, Alexei~A. Efros, and Sergey Levine,
\newblock ``Conditional networks for few-shot semantic segmentation,''
\newblock in {\em ICLR}, 2018.

\bibitem{raza2019weakly}
Hasnain Raza, Mahdyar Ravanbakhsh, Tassilo Klein, and Moin Nabi,
\newblock ``Weakly supervised one shot segmentation,''
\newblock in {\em ICCVW}, 2019.

\bibitem{siam2020weakly}
Mennatullah Siam, Naren Doraiswamy, Boris~N Oreshkin, Hengshuai Yao, and Martin Jagersand,
\newblock ``Weakly supervised few-shot object segmentation using co-attention with visual and semantic embeddings,''
\newblock {\em arXiv preprint arXiv:2001.09540}, 2020.

\bibitem{lee2022pixel}
Yuan-Hao Lee, Fu-En Yang, and Yu-Chiang~Frank Wang,
\newblock ``A pixel-level meta-learner for weakly supervised few-shot semantic segmentation,''
\newblock in {\em WACV}, 2022, pp. 2170--2180.

\bibitem{selvaraju2017grad}
Ramprasaath~R Selvaraju, Michael Cogswell, Abhishek Das, Ramakrishna Vedantam, Devi Parikh, and Dhruv Batra,
\newblock ``Grad-cam: Visual explanations from deep networks via gradient-based localization,''
\newblock in {\em ICCV}, 2017, pp. 618--626.

\bibitem{wangiterative}
Haohan Wang, Liang Liu, Wuhao Zhang, Jiangning Zhang, Zhenye Gan, Yabiao Wang, Chengjie Wang, and Haoqian Wang,
\newblock ``Iterative few-shot semantic segmentation from image label text,''
\newblock in {\em IJCAI}, 2022.

\bibitem{radford2021learning}
Alec Radford, Jong~Wook Kim, Chris Hallacy, Aditya Ramesh, Gabriel Goh, Sandhini Agarwal, Girish Sastry, Amanda Askell, Pamela Mishkin, Jack Clark, et~al.,
\newblock ``Learning transferable visual models from natural language supervision,''
\newblock in {\em ICML}, 2021, pp. 8748--8763.

\bibitem{zhou2022extract}
Chong Zhou, Chen~Change Loy, and Bo~Dai,
\newblock ``Extract free dense labels from clip,''
\newblock in {\em ECCV}, 2022, pp. 696--712.

\bibitem{tian2020prior}
Zhuotao Tian, Hengshuang Zhao, Michelle Shu, Zhicheng Yang, Ruiyu Li, and Jiaya Jia,
\newblock ``Prior guided feature enrichment network for few-shot segmentation,''
\newblock {\em IEEE TPAMI}, 2020.

\bibitem{lang2022learning}
Chunbo Lang, Gong Cheng, Binfei Tu, and Junwei Han,
\newblock ``Learning what not to segment: A new perspective on few-shot segmentation,''
\newblock in {\em CVPR}, 2022, pp. 8057--8067.

\bibitem{min2021hypercorrelation}
Juhong Min, Dahyun Kang, and Minsu Cho,
\newblock ``Hypercorrelation squeeze for few-shot segmentation,''
\newblock in {\em ICCV}, 2021, pp. 6941--6952.

\bibitem{iqbal2022msanet}
Ehtesham Iqbal, Sirojbek Safarov, and Seongdeok Bang,
\newblock ``Msanet: Multi-similarity and attention guidance for boosting few-shot segmentation,''
\newblock {\em arXiv preprint arXiv:2206.09667}, 2022.

\bibitem{ni2022expanding}
Bolin Ni, Houwen Peng, Minghao Chen, Songyang Zhang, Gaofeng Meng, Jianlong Fu, Shiming Xiang, and Haibin Ling,
\newblock ``Expanding language-image pretrained models for general video recognition,''
\newblock in {\em ECCV}, 2022, pp. 1--18.

\bibitem{gal2022stylegan}
Rinon Gal, Or~Patashnik, Haggai Maron, Amit~H Bermano, Gal Chechik, and Daniel Cohen-Or,
\newblock ``Stylegan-nada: Clip-guided domain adaptation of image generators,''
\newblock {\em ACM Transactions on Graphics (TOG)}, vol. 41, no. 4, pp. 1--13, 2022.

\bibitem{luo2022clip4clip}
Huaishao Luo, Lei Ji, Ming Zhong, Yang Chen, Wen Lei, Nan Duan, and Tianrui Li,
\newblock ``Clip4clip: An empirical study of clip for end to end video clip retrieval and captioning,''
\newblock {\em Neurocomputing}, vol. 508, pp. 293--304, 2022.

\bibitem{gu2021open}
Xiuye Gu, Tsung-Yi Lin, Weicheng Kuo, and Yin Cui,
\newblock ``Open-vocabulary detection via vision and language knowledge distillation,''
\newblock {\em arXiv preprint arXiv:2104.13921}, 2021.

\bibitem{ding2022decoupling}
Jian Ding, Nan Xue, Gui-Song Xia, and Dengxin Dai,
\newblock ``Decoupling zero-shot semantic segmentation,''
\newblock in {\em CVPR}, 2022, pp. 11583--11592.

\bibitem{pourpanah2022review}
Farhad Pourpanah, Moloud Abdar, Yuxuan Luo, Xinlei Zhou, Ran Wang, Chee~Peng Lim, Xi-Zhao Wang, and QM~Jonathan Wu,
\newblock ``A review of generalized zero-shot learning methods,''
\newblock {\em IEEE TPAMI}, 2022.

\bibitem{rao2022denseclip}
Yongming Rao, Wenliang Zhao, Guangyi Chen, Yansong Tang, Zheng Zhu, Guan Huang, Jie Zhou, and Jiwen Lu,
\newblock ``Denseclip: Language-guided dense prediction with context-aware prompting,''
\newblock in {\em CVPR}, 2022, pp. 18082--18091.

\bibitem{shaban2017one}
Amirreza Shaban, Shray Bansal, Zhen Liu, Irfan Essa, and Byron Boots,
\newblock ``One-shot learning for semantic segmentation,''
\newblock {\em arXiv preprint arXiv:1709.03410}, 2017.

\bibitem{li2021adaptive}
Gen Li, Varun Jampani, Laura Sevilla-Lara, Deqing Sun, Jonghyun Kim, and Joongkyu Kim,
\newblock ``Adaptive prototype learning and allocation for few-shot segmentation,''
\newblock in {\em CVPR}, 2021, pp. 8334--8343.

\bibitem{lin2014microsoft}
Tsung-Yi Lin, Michael Maire, Serge Belongie, James Hays, Pietro Perona, Deva Ramanan, Piotr Doll{\'a}r, and C~Lawrence Zitnick,
\newblock ``Microsoft coco: Common objects in context,''
\newblock in {\em ECCV}, 2014, pp. 740--755.

\end{thebibliography}

\end{document}